\documentclass[10pt,twocolumn,letterpaper]{article}

\usepackage{cvpr}
\usepackage{times}
\usepackage{epsfig}
\usepackage{graphicx}
\usepackage{amsmath}
\usepackage{amssymb}
\usepackage{graphics}
\usepackage{subfigure}
\usepackage{epstopdf}
\usepackage{array}
\usepackage{multirow}
\usepackage{color}
\usepackage{array}
\usepackage{booktabs}
\usepackage{comment}

\usepackage[breaklinks=true,bookmarks=false]{hyperref}

\cvprfinalcopy 

\def\cvprPaperID{3} 

\ifcvprfinal\fi
\begin{document}

\title{DPW-SDNet: Dual Pixel-Wavelet Domain Deep CNNs \\for Soft Decoding of JPEG-Compressed Images}

\author{Honggang Chen\\
Sichuan University\\
{\tt\small honggang\_chen@yeah.net}
\and
\and
Xiaohai He\\
Sichuan University\\
{\tt\small hxh@scu.edu.cn}
\and
\and
Linbo Qing\\
Sichuan University\\
{\tt\small qing\_lb@scu.edu.cn}
\and
Shuhua Xiong\\
Sichuan University\\
{\tt\small xiongsh@scu.edu.cn}
\and
Truong Q. Nguyen\\
UC San Diego\\
{\tt\small tqn001@eng.ucsd.edu}
}

\maketitle
\thispagestyle{empty}

\begin{abstract}
JPEG is one of the widely used lossy compression methods. JPEG-compressed images usually suffer from compression artifacts including blocking and blurring, especially at low bit-rates. Soft decoding is an effective solution to improve the quality of compressed images without changing codec or introducing extra coding bits. Inspired by the excellent performance of the deep convolutional neural networks (CNNs) on both low-level and high-level computer vision problems, we develop a dual pixel-wavelet domain deep CNNs-based soft decoding network for JPEG-compressed images, namely DPW-SDNet. The pixel domain deep network takes the four downsampled versions of the compressed image to form a 4-channel input and outputs a pixel domain prediction, while the wavelet domain deep network uses the 1-level discrete wavelet transformation (DWT) coefficients to form a 4-channel input to produce a DWT domain prediction. The pixel domain and wavelet domain estimates are combined to generate the final soft decoded result. Experimental results demonstrate the superiority of the proposed DPW-SDNet over several state-of-the-art compression artifacts reduction algorithms.
\end{abstract}

\section{Introduction}

\begin{figure}[!tb]
    \centering
    \subfigure[]{
    \includegraphics[width=4cm]{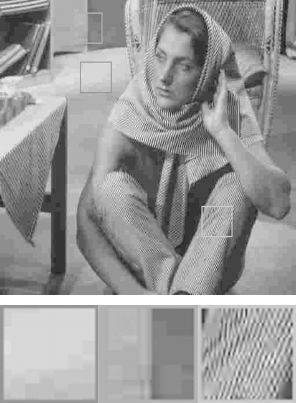}}
    \subfigure[]{
    \includegraphics[width=4cm]{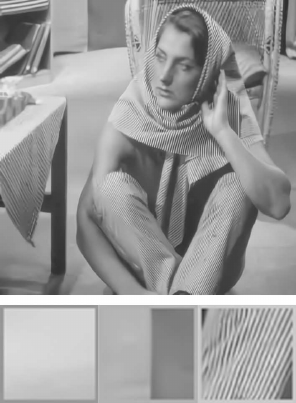}}
    \caption{Illustrations of compression artifacts and soft decoding. (a) JPEG-compressed image in the case of QF = 10 (PSNR = 25.79 dB, SSIM = 0.7621, PSNR-B = 23.48 dB); (b) Soft decoded result of (a) using the developed DPW-SDNet (PSNR = {\textcolor{red}{28.22}} dB, SSIM = {\textcolor{red}{0.8376}}, PSNR-B = {\textcolor{red}{27.84}} dB).}
    \label{fig.1}
\end{figure}

The number of devices with high-resolution camera increases significantly over the last few years, with the introduction of smart phones and IoT (Internet of Things) devices. Limited by the transmission bandwidth and storage capacity, these images and videos are compressed. As shown in Fig. \ref{fig.1}, compressed images usually suffer from compression artifacts due to the information loss in the lossy compression process, especially at low bit-rates. In addition to poor perceptual quality, compression artifacts also reduce the accuracy of other processing steps such as object detection and classification. Therefore, it is necessary to improve the quality of compressed images. This paper focuses on the soft decoding of JPEG images due to the fact that the JPEG is one of the commonly used compression standards for still images.

In recent years, many works investigate the restoration of JPEG images, aiming to remove compression artifacts and enhance the perceptual quality and objective assessment scores. In literature, the restoration procedure is usually referred to as soft decoding \cite{Liu2017Random, Liu2016Data}, deblocking \cite{ Li2017Iterative, Zhang2016CONCOLOR}, or compression artifacts reduction \cite{Dong2015Compression, guo2017one}. In this paper, we use these terms interchangeably. Inspired by the excellent performance of the deep convolutional neural networks (CNNs) on various computer vision problems, we propose a dual pixel-wavelet domain deep CNNs-based soft decoding network for JPEG-compressed images, namely DPW-SDNet. From Fig. \ref{fig.1} that illustrates a restored image by the proposed DPW-SDNet, we can observe that most of the compression artifacts are removed and some missing textures are recovered. Overall, the main contribution of this work is a dual-branch deep CNN that can reduce compression artifacts in both the pixel domain and wavelet domain. More specifically, our contributions are two folds:
\begin{itemize}

\item We develop an effective and efficient soft decoding method for JPEG-compressed images using dual pixel-wavelet domain deep CNNs. The combination of the pixel domain and wavelet domain predictions leads to
    better soft decoding performance.

\item We reshape the compressed image and its 1-level discrete wavelet transformation (DWT) coefficients into two tensors with smaller size, which are used as the inputs to the pixel and wavelet sub-networks, respectively. By performing soft decoding on two smaller tensors, the DPW-SDNet achieves state-of-the-art performance while maintaining efficiency.

\end{itemize}

The rest of this paper is organized as follows. We describe the related work in the next section. The proposed soft decoding algorithm is presented in Section 3. Experiments are shown in Section 4. Finally, Section 5 concludes this paper.

\section{Related Work}

Let ${\bf{X}}$ and ${\bf{Y}}$ be the original uncompressed image and the corresponding JPEG-compressed version, respectively. Given the compressed image ${\bf{Y}}$, the goal of soft decoding is to produce an estimate that is as close as possible to the original image ${\bf{X}}$. Existing methods for soft decoding of JPEG-compressed images can be roughly split into three categories: enhancement-based, restoration-based, and learning-based methods.

The enhancement-based methods usually remove compression artifacts via performing pixel domain or transform domain filtering. For instance, Foi et al. \cite{Foi2007Pointwise} proposed a shape-adaptive discrete cosine transformation (DCT)-based image filtering, yielding excellent performance on deblocking and deringing of compressed images.  Zhai et al. \cite{zhai2008efficient1} proposed to reduce blocking artifacts via postfiltering in shifted windows of image blocks. In \cite{zhai2008efficient2}, the authors developed an efficient artifacts reduction algorithm through joint DCT domain and spatial domain processing. Yoo et al. \cite{yoo2014post} proposed an inter-block correlation-based blocking artifacts reduction framework, in which the artifacts in flat regions and edge regions were removed using different strategies.

Compression artifacts reduction is formulated as an ill-posed inverse problem for the restoration-based soft decoding methods, where the prior knowledge about high-quality images, compression algorithms, and compression parameters is used to assist the restoration process \cite{Chang2014Reducing, Dar2016Postprocessing, Hu2016Graph, Li2017Iterative,  Liu2017Random, Liu2016Data, mu2016adaptive, ren2013image, Sun2007Postprocessing, zhang2015image, Zhang2016CONCOLOR, Zhang2016Low, Zhang2013Compression,  zhao2017Reducing}. For instance, in \cite{Sun2007Postprocessing}, the original image and compression distortion were modeled as a high-order Markov random field and spatially correlated Gaussian noise, respectively.
Non-local self-similarity property was widely used in deblocking algorithms. In general, the low-rank \cite{Li2017Iterative, ren2013image, Zhang2016CONCOLOR, Zhang2016Low} and group sparse representation \cite{zhang2015image, zhao2017Reducing} were applied to model this property.
In \cite{Chang2014Reducing, Liu2017Random, Liu2016Data, mu2016adaptive, zhang2015image, zhao2017Reducing}, sparsity was utilized as an image prior to regularize the restored image.
The graph model was used in the deblocking methods proposed in \cite{Hu2016Graph} and \cite{Liu2017Random}.
In some works \cite{ Liu2017Random, Liu2016Data, Zhang2016CONCOLOR,  Zhang2016Low, zhao2017Reducing}, the quantization constraint on DCT coefficients was applied to  restrain the resultant image.
In particular, Dar et al. \cite{Dar2016Postprocessing} designed a sequential denoising-based soft decoding algorithm, where the existing state-of-the-art denoising method was used to construct a regularization.  On the whole, most of the restoration-based soft decoding methods are time-consuming to some extent due to the complex optimization process.

Recently, excellent results were obtained by deep learning-based approaches \cite{Cavigelli2017CAS, chen2017trainable, Dong2015Compression, Galteri_2017_ICCV, Guo2016Building, guo2017one, Li2017An, Wang2016D3, Zhang2017Beyond}. Dong et al. \cite{Dong2015Compression} developed a shallow CNN for compression artifacts reduction on the basis of the network for super-resolution \cite{dong2014learning}. The authors of \cite{Dong2015Compression} found that it is hard to train a network beyond four layers in low-level vision tasks. To address this issue, Kim et al. \cite{kim2016accurate} introduced the residual learning technique and designed a very deep network of twenty layers for single image super-resolution. In \cite{Zhang2017Beyond}, Zhang et al. presented a very deep network via incorporating the residual learning and batch normalization for a series of general image denoising problems, including denoising, super-resolution, and deblocking.  Li et al. \cite{Li2017An} combined the skip connection and residual learning to ease the network training process. Cavigelli et al. \cite{Cavigelli2017CAS} developed a deep compression artifacts reduction network with a multi-scale loss function. In \cite{chen2017trainable}, Chen and Pock proposed a trainable nonlinear reaction diffusion model for efficient image restoration. Inspired by the success of the dual DCT-pixel domain sparse coding \cite{Liu2016Data}, the authors of \cite{Guo2016Building} and \cite{Wang2016D3} designed dual-domain networks for the deblocking of JPEG images. More recently, some works aim to improve the perceptual quality of compressed images \cite{Galteri_2017_ICCV, guo2017one}. Overall, deep learning-based approaches show obvious superiority over conventional soft decoding methods in terms of both the restoration performance and running time \footnote{\label{footnote0} In general, the deep learning-based image restoration approaches are time-consuming in model training phase but efficient in testing phase. In this paper, the running time refers to the time cost in testing phase only.}.

\begin{figure*}[!tb]
    \centering
    \includegraphics[width=12cm]{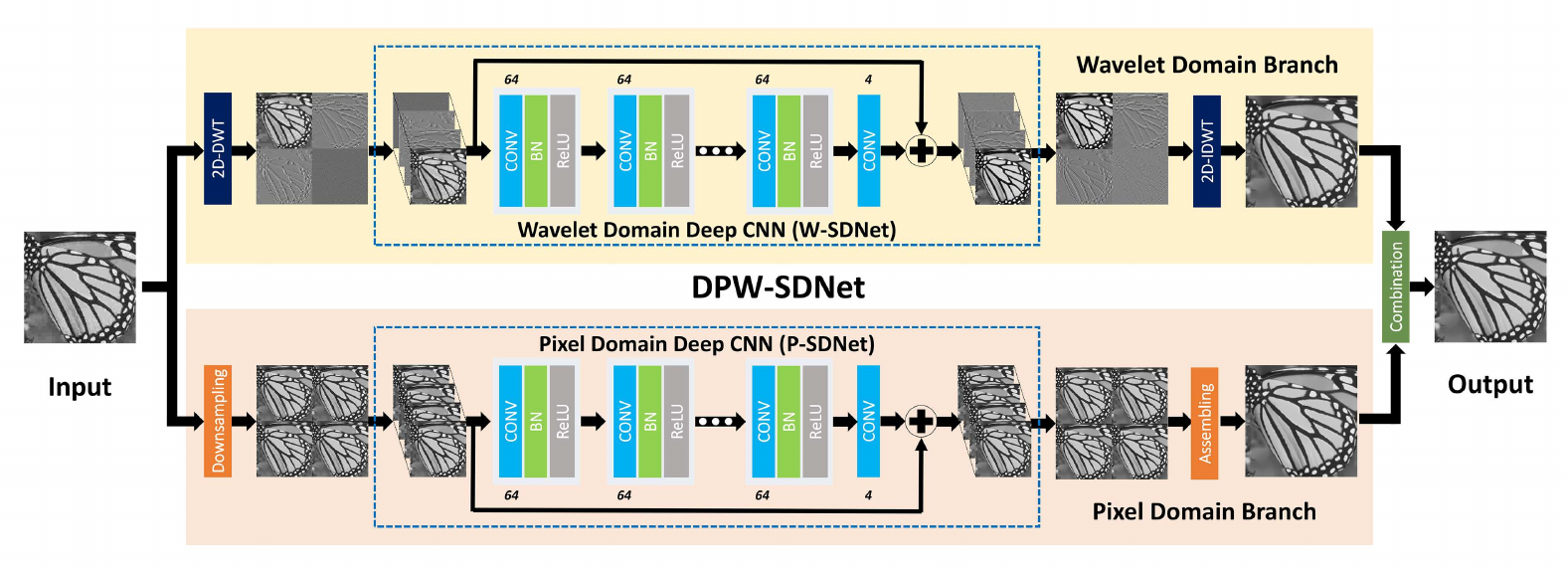}
    \caption{Flowchart of the proposed DPW-SDNet. The DPW-SDNet reduces compression artifacts in dual pixel-wavelet domain. The depths of the P-SDNet and W-SDNet are set to $20$. The number next to each convolutional layer represents the number of kernels, and all of the convolutional layers in DPW-SDNet have the same kernel size of $3 \times 3$.}
    \label{fig.2}
\end{figure*}

\begin{figure}[!tb]
    \centering
    \subfigure[]{
    \includegraphics[width=2.5cm]{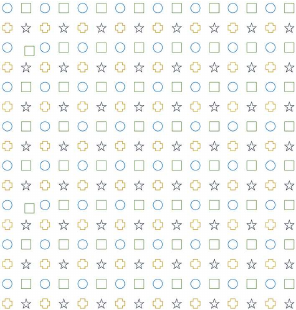}}
\hspace{5pt}
    \subfigure[\hspace{-5pt}]{
    \includegraphics[width=1.25cm]{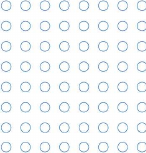}}
    \hspace{5pt}
   \subfigure[\hspace{-5pt}]{
    \includegraphics[width=1.25cm]{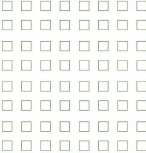}}\\
     \hspace{20pt}
    \subfigure[\hspace{-5pt}]{
    \includegraphics[width=1.25cm]{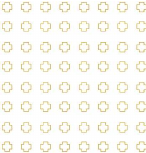}}
     \hspace{25pt}
   \subfigure[\hspace{-5pt}]{
    \includegraphics[width=1.25cm]{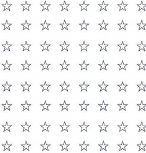}}
     \hspace{0pt}
    \subfigure[\hspace{-5pt}]{
    \includegraphics[width=1.6cm]{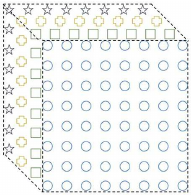}}
    \caption{Illustration of the reversible downsampling process used in the pixel domain soft decoding branch. (a) The input image (size: $m \times n$, here $m=n=16$); (b)-(e) Different downsampled versions of (a) (size: $\frac{m}{2} \times \frac{n}{2}$); (f) The tensor composed of (b)-(e) (size: $\frac{m}{2} \times \frac{n}{2} \times 4$). Note that this downsampling process is reversible. }
    \label{fig.3}
\end{figure}

Inspired by the success of the wavelet domain networks for super-resolution \cite{guo2017deep, huang2017wavelet}, we present a dual pixel-wavelet domain deep CNN for the soft decoding of JPEG-compressed images in this paper. The proposed DPW-SDNet is different from previous deep learning-based soft decoding algorithms in the following aspects: 1) The DPW-SDNet consists of two parallel branches that perform restoration in the pixel domain and wavelet domain, respectively. 2) The DPW-SDNet takes two tensors as the network inputs rather than the original compressed image and DWT coefficients. Experiments show that the DPW-SDNet achieves competitive restoration performance and execution speed on JPEG-compressed images. Moreover, the extensions of the proposed DPW-SDNet to other compression standards are straightforward.

\section{Proposed DPW-SDNet}

As outlined in Fig. \ref{fig.2}, the proposed DPW-SDNet composes of two parallel branches: the pixel domain soft decoding branch and the wavelet domain soft decoding branch. The network in the pixel domain branch (namely P-SDNet) removes compression artifacts in pixel domain directly, while the network in the wavelet domain branch (namely W-SDNet) performs restoration in wavelet domain. The pixel domain and wavelet domain estimates are combined to generate the final soft decoded result. Note that we do not directly use the original compressed image and its DWT sub-bands as the inputs of the two sub-networks. In the following sections, more details about the DPW-SDNet are presented. For convenience, we assume that the input ${\bf{Y}}$ is a gray-scaled image of size $m \times n$ where $m,n$ are both even.

\subsection{The Pixel Domain Branch}

In the pixel domain branch (shown in the bottom half of Fig. \ref{fig.2}), first the compressed image ${\bf{Y}}$  is downsampled to generate four downsampled sub-images of size $\frac{m}{2} \times \frac{n}{2}$. Since we have to recover an image that has the same size with the input, a reversible downsampling strategy is used in this process as \cite{zhang2017ffdnet}. Fig. \ref{fig.3} illustrates the reversible downsampling process. Given ${\bf{Y}}$, the pixels located at $(2i+1,2j+1)$, $(2i+1,2j+2)$, $(2i+2,2j+1)$, and $(2i+2,2j+2)$ ($i = 0,1,2, \cdots ,\frac{m}{2} - 1$, $j = 0,1,2, \cdots ,\frac{n}{2} - 1$) are respectively sampled to form four different sub-images, which are concatenated to constitute a tensor of size $\frac{m}{2} \times \frac{n}{2} \times 4$. Then, the tensor is fed into the pixel domain deep CNN. At least two benefits can be achieved by using a smaller tensor as the input of a deep CNN. First, a smaller input means lower computational complexity. In addition, working on the downsampled images can enlarge the receptive field, which is beneficial to restoration process.

For convenience, we name the pixel domain deep CNN P-SDNet. The input and output of the P-SDNet are tensors.  The ${D}$-layer P-SDNet consists of two kinds of blocks. The first ${(D-1)}$ blocks are ``CONV+BN+ReLU", and the last block only includes a convolutional layer. Note that the abbreviation ``CONV" represents a convolutional layer, ``BN" denotes the batch normalization \cite{ioffe2015batch}, and ``ReLU" represents the rectified linear unit \cite{krizhevsky2012imagenet}. The kernel number of each convolutional layer is set to $64$ except the last layer that outputs a $4$-channel residual image. The kernel size of each convolutional layers is set to $3 \times 3$.  In each layer, the zero padding strategy is adopted to keep all feature maps having the same size. Since the input and output of the P-SDNet are very similar, we adopt the residual learning \cite{he2016deep} for stable and fast training. Hence, the training loss function of the P-SDNet is defined as
\begin{equation}
\label{eq.1}
{L_P}({\Theta _P}) = \frac{1}{2N}\sum\limits_{i = 1}^N {{{\left\| {(f_p({\bf{y}}_i^{pt};{\Theta _P})+{\bf{y}}_i^{pt}) -{\bf{x}}_i^{pt}} \right\|}^2}}
\end{equation}
where the $\Theta _P$ represents all parameters in P-SDNet, $f_p({\bf{y}}_i^{pt};{\Theta _P})$ is the predicted residual component, and $\{ ({\bf{y}}_i^{pt},{\bf{x}}_i^{pt})\} _{i = 1}^N$ denotes $N$ compressed-clean tensor pairs in the pixel domain.

Finally, the four feature maps in the output of P-SDNet are assembled according to the inverse process of the downsampling procedure to form the pixel domain estimate.

\subsection{The Wavelet Domain Branch}

The framework of the wavelet domain branch is similar to the pixel domain branch. Given a compressed image ${\bf{Y}}$, we first conduct the 1-level 2-dimensional discrete wavelet transformation (2D-DWT) and obtain its four wavelet sub-bands coefficients. The size of each sub-band is $\frac{m}{2} \times \frac{n}{2}$. Similarly, the four wavelet sub-bands are concatenated to constitute a tensor of size $\frac{m}{2} \times \frac{n}{2} \times 4$, which is used as the input of the wavelet domain deep CNN, namely W-SDNet. By concatenating four wavelet sub-bands, the information in different sub-bands can be fused while keeping the consistency among them. Moreover, the computational cost can be reduced.

The architecture of the W-SDNet is set to be the same as the P-SDNet, including the network depth, number of kernels, and kernel size. Therefore, we do not introduce the W-SDNet in details to avoid redundancy. The main difference between the two sub-networks is that the W-SDNet predicts wavelet coefficients residual while the P-SDNet predicts pixel intensity residual. Correspondingly, the training loss function of the W-SDNet is defined as
\begin{equation}
\label{eq.2}
{L_W}({\Theta _W}) = \frac{1}{2N}\sum\limits_{i = 1}^N {{{\left\|( {f_w({\bf{y}}_i^{wt};{\Theta _W})+{\bf{y}}_i^{wt}) - {\bf{x}}_i^{wt}} \right\|}^2}}
\end{equation}
where the $\Theta _W$ represents all parameters in W-SDNet, $f_w({\bf{y}}_i^{wt};{\Theta _W})$ is the predicted residual component, and $\{ ({\bf{y}}_i^{wt},{\bf{x}}_i^{wt})\} _{i = 1}^N$ denotes $N$ compressed-clean tensor pairs in the wavelet domain.

The four feature maps in the output of W-SDNet are the wavelet sub-bands of the soft decoded image. Therefore, the 2-dimensional inverse discrete wavelet transformation (2D-IDWT) is performed on these coefficients to produce the wavelet domain estimate.

\subsection{The Combination of the Dual-Branch}

As mentioned above, the pixel domain and wavelet domain branches both produce a soft decoded version of the input image. Since the two predictions are generated in different spaces, they have their respective characteristics. Hence, combining them should improve the restoration performance further. There are many ways to fuse the two intermediate results. For example, we can design a network with a 2-channel input and a 1-channel output to combine them. Considering the computational complexity, the two estimates derived from the dual-domain are simply equally weighted to generate the final output in this work.

\section{Experiments}

In this section, we first introduce some implementation details, followed by experimental results.

\subsection{Implementation Details}

\textbf{Training Data}: The publicly available imageset BSDS500 \footnote{\label{footnote1}Available: https://www2.eecs.berkeley.edu/Research/Projects/CS/\\vision/grouping/resources.html} is used to train the DPW-SDNet. We adopt the data augmentation (rotation and downsampling) to generate more training images. For the P-SDNet, we extract training sample pairs from original images and the corresponding compressed images. Correspondingly, the 2D-DWT coefficients of the original images and compressed images are used to generate training sample pairs for the W-SDNet. We generate $N = 523,968$ training sample pairs for each sub-network, and the size of each sample is set to $31 \times 31 \times 4$.

\textbf{Training Parameters}: We use the Caffe package \cite{jia2014caffe} to implement the proposed network, and the depths of P-SDNet and W-SDNet are set to $20$ ($D = 20$). The stochastic gradient descent algorithm is adopted to optimize our networks. The batch size, weight decay, momentum are set to $64$, $0.0001$, and $0.9$, respectively. The initial learning rate is set to $0.1$, and it decreases by a factor of $10$ every $10$ epochs. The maximum number of iterations is set to $300,000$ for both the pixel domain and wavelet domain sub-networks.

\begin{table*}[!tb]
\centering
\caption{Average PSNR (dB)/SSIM/PSNR-B (dB) scores of different soft decoding algorithms on Classic$5$ and LIVE$1$. The best and the second-best scores are highlighted in {\color{red}{red}} and {\color{blue}{blue}}, respectively. }
\begin{tabular}{|p{1cm}<{\centering}|p{2.7cm}<{\centering}|p{2.75cm}<{\centering}|p{2.75cm}<{\centering}|p{2.75cm}<{\centering}|p{2.75cm}<{\centering}|}
\hline
\multicolumn{2}{|c|}{QF} & 10 & 20 & 30 & 40\\
\hline
\multirow{7}{*}{Classic5}
& JPEG  & 27.82/0.7595/25.21  &  30.12/0.8344/27.50 & 31.48/0.8666/28.94 & 32.43/0.8849/29.92\\
\cline{2-6}
& CONCOLOR \cite{Zhang2016CONCOLOR}  & 29.24/0.7963/{\textcolor{blue}{29.14}} & 31.38/0.8541/31.18 & 32.70/0.8809/{\textcolor{blue}{32.50}} & 33.60/0.8961/{\textcolor{red}{33.36}} \\
\cline{2-6}
& D2SD \cite{Liu2016Data}  & 29.21/0.7960/28.87 & 31.47/0.8551/31.15 & 32.79/0.8813/32.40 &       33.66/0.8962/33.20 \\
\cline{2-6}
& ARCNN \cite{Dong2015Compression}  & 29.05/0.7929/28.78 & 31.16/0.8517/30.60 & 32.52/0.8806/32.00 & 33.33/0.8953/32.81 \\
\cline{2-6}
& TNRD \cite{chen2017trainable} & 29.28/0.7992/29.04 & 31.47/0.8576/31.05 & 32.78/0.8837/32.24 & - \\
\cline{2-6}
& DnCNN-3 \cite{Zhang2017Beyond} & {\textcolor{blue}{29.40}}/{\textcolor{blue}{0.8026}}/29.13  & {\textcolor{blue}{31.63}}/{\textcolor{blue}{0.8610}}/{\textcolor{blue}{31.19}}  & {\textcolor{blue}{32.90}}/{\textcolor{blue}{0.8860}}/32.36 & {\textcolor{blue}{33.77}}/{\textcolor{blue}{0.9003}}/33.20 \\
\cline{2-6}
& DPW-SDNet  & {\textcolor{red}{29.74}}/{\textcolor{red}{0.8124}}/{\textcolor{red}{29.37}} & {\textcolor{red}{31.95}}/{\textcolor{red}{0.8663}}/{\textcolor{red}{31.42}}& {\textcolor{red}{33.22}}/{\textcolor{red}{0.8903}}/{\textcolor{red}{32.51}} & {\textcolor{red}{34.07}}/{\textcolor{red}{0.9039}}/{\textcolor{blue}{33.24}}\\
\hline
\multicolumn{6}{|c|}{}\\
\hline
\multirow{7}{*}{LIVE1}
& JPEG  & 27.77/0.7730/25.34 & 30.08/0.8512/27.57 & 31.41/0.8852/28.93 & 32.36/0.9041/29.96 \\
\cline{2-6}
& CONCOLOR \cite{Zhang2016CONCOLOR}  & 28.87/0.8018/28.76  & 31.08/0.8681/30.90 & 32.42/0.8985/32.16 & 33.39/0.9157/33.07\\
\cline{2-6}
& D2SD \cite{Liu2016Data}  & 28.83/0.8023/28.54 & 31.08/0.8690/30.80 & 32.41/0.8987/32.10 & 33.37/0.9156/33.06 \\
\cline{2-6}
& ARCNN \cite{Dong2015Compression}  & 29.04/0.8076/28.77 & 31.31/0.8733/30.79 & 32.73/0.9043/32.22 & 33.63/0.9198/33.14 \\
\cline{2-6}
& TNRD \cite{chen2017trainable} & 29.14/0.8111/28.88 & 31.46/0.8769/31.04 & 32.84/0.9059/32.28 & -\\
\cline{2-6}
& DnCNN-3 \cite{Zhang2017Beyond} & {\textcolor{blue}{29.19}}/{\textcolor{blue}{0.8123}}/{\textcolor{blue}{28.91}} & {\textcolor{blue}{31.59}}/{\textcolor{blue}{0.8802}}/{\textcolor{blue}{31.08}} & {\textcolor{blue}{32.99}}/{\textcolor{blue}{0.9090}}/{\textcolor{blue}{32.35}} &   {\textcolor{blue}{33.96}}/{\textcolor{blue}{0.9247}}/{\textcolor{blue}{33.29}} \\
\cline{2-6}
& DPW-SDNet  & {\textcolor{red}{29.53}}/{\textcolor{red}{0.8210}}/{\textcolor{red}{29.13}} &{\textcolor{red}{31.90}}/{\textcolor{red}{0.8854}}/{\textcolor{red}{31.27}}  & {\textcolor{red}{33.31}}/{\textcolor{red}{0.9130}}/{\textcolor{red}{32.52}}& {\textcolor{red}{34.30}}/{\textcolor{red}{0.9282}}/{\textcolor{red}{33.44}}\\
\hline
\end{tabular}
\label{table.1}
\end{table*}
\begin{figure*}[!tb]
    \centering
    \subfigure[Original image]{
    \includegraphics[width=4cm]{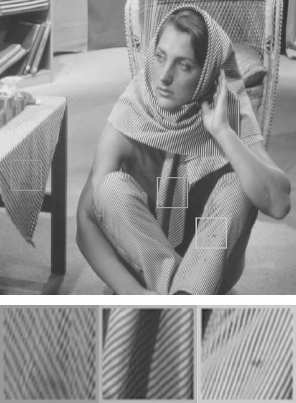}}
    \subfigure[JPEG]{
   \includegraphics[width=4cm]{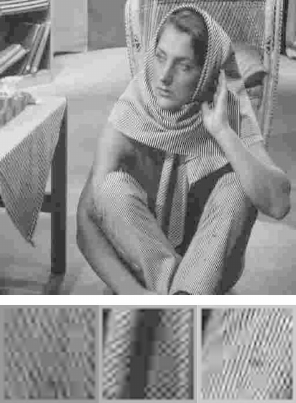}}
    \subfigure[CONCOLOR \cite{Zhang2016CONCOLOR}]{
   \includegraphics[width=4cm]{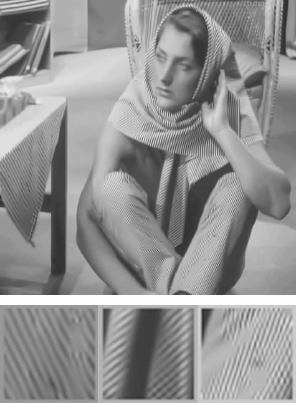}}
    \subfigure[D2SD \cite{Liu2016Data}]{
   \includegraphics[width=4cm]{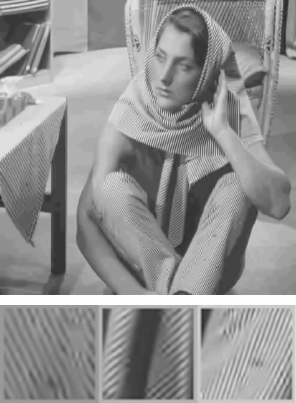}}\\
    \subfigure[ARCNN \cite{Dong2015Compression}]{
   \includegraphics[width=4cm]{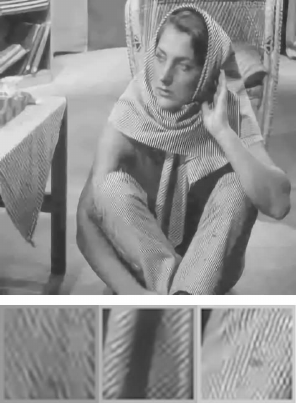}}
    \subfigure[TNRD \cite{chen2017trainable}]{
   \includegraphics[width=4cm]{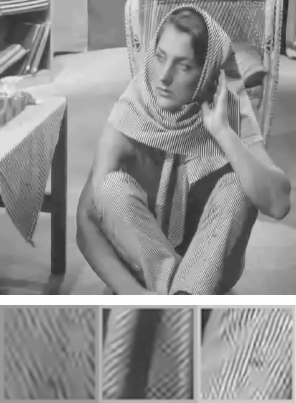}}
    \subfigure[DnCNN-3 \cite{Zhang2017Beyond}]{
    \includegraphics[width=4cm]{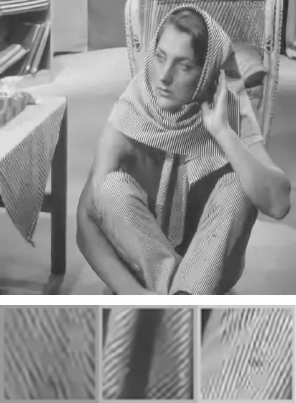}}
    \subfigure[Proposed DPW-SDNet]{
    \includegraphics[width=4cm]{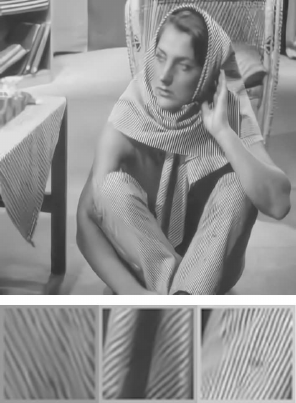}}
    \caption{Visual quality comparison of different soft decoding methods on \emph{Barbara} in the case of QF = 10. (a) Original image (PSNR (dB), SSIM, PSNR-B (dB)); (b) JPEG (25.79, 0.7621, 23.48); (c) CONCOLOR \cite{Zhang2016CONCOLOR} (27.73, {\textcolor{blue}{0.8216}}, 27.63); (d) D2SD \cite{Liu2016Data} ({\textcolor{blue}{27.93}}, 0.8214, {\textcolor{blue}{27.64}}); (e) ARCNN \cite{Dong2015Compression} (26.92, 0.7967, 26.75); (f) TNRD \cite{chen2017trainable} (27.24, 0.8099, 27.13); (g) DnCNN-3 \cite{Zhang2017Beyond} (27.58, 0.8161, 27.29); (h) Proposed DPW-SDNet ({\textcolor{red}{28.22}}, {\textcolor{red}{0.8376}}, {\textcolor{red}{27.84}}).}
    \label{fig.4}
\end{figure*}

\begin{figure*}[!tb]
    \centering
    \subfigure[Original image]{
    \includegraphics[width=4cm]{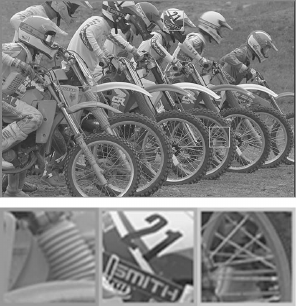}}
    \subfigure[JPEG]{
   \includegraphics[width=4cm]{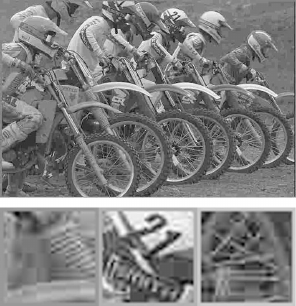}}
    \subfigure[CONCOLOR \cite{Zhang2016CONCOLOR}]{
   \includegraphics[width=4cm]{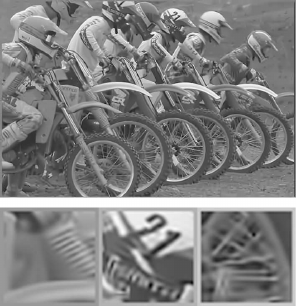}}
    \subfigure[D2SD \cite{Liu2016Data}]{
   \includegraphics[width=4cm]{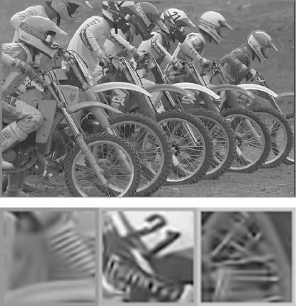}}\\
    \subfigure[ARCNN \cite{Dong2015Compression}]{
   \includegraphics[width=4cm]{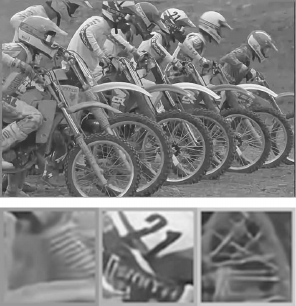}}
    \subfigure[TNRD \cite{chen2017trainable}]{
   \includegraphics[width=4cm]{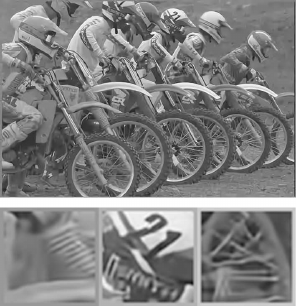}}
    \subfigure[DnCNN-3 \cite{Zhang2017Beyond}]{
    \includegraphics[width=4cm]{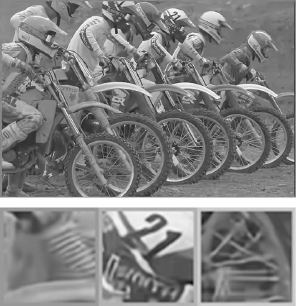}}
    \subfigure[Proposed DPW-SDNet]{
    \includegraphics[width=4cm]{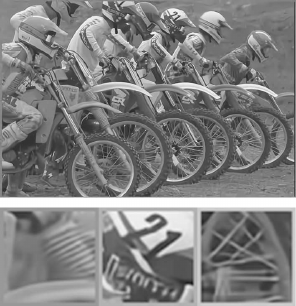}}
    \caption{Visual quality comparison of different soft decoding methods on \emph{Bike} in the case of QF = 10. (a) Original image (PSNR (dB), SSIM, PSNR-B (dB)); (b) JPEG (25.77, 0.7417, 23.02); (c) CONCOLOR \cite{Zhang2016CONCOLOR} (27.00, 0.7801, 27.00); (d) D2SD \cite{Liu2016Data} (27.11, 0.7859, 26.97); (e) ARCNN \cite{Dong2015Compression} (27.41, 0.7924, 27.11); (f) TNRD \cite{chen2017trainable} (27.54, 0.7971, 27.22); (g) DnCNN-3 \cite{Zhang2017Beyond} ({\textcolor{blue}{27.59}}, {\textcolor{blue}{0.7999}}, {\textcolor{blue}{27.28}}); (h) Proposed DPW-SDNet ({\textcolor{red}{28.04}}, {\textcolor{red}{0.8133}}, {\textcolor{red}{27.58}}).}
    \label{fig.5}
\end{figure*}

\begin{figure*}[!tb]
    \centering
    \subfigure[Original image]{
    \includegraphics[width=4cm]{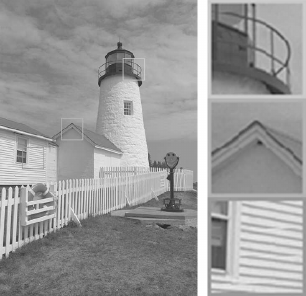}}
    \subfigure[JPEG]{
   \includegraphics[width=4cm]{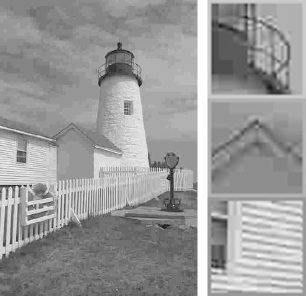}}
    \subfigure[CONCOLOR \cite{Zhang2016CONCOLOR}]{
   \includegraphics[width=4cm]{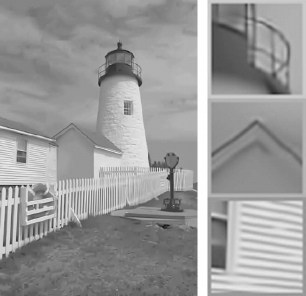}}
    \subfigure[D2SD \cite{Liu2016Data}]{
   \includegraphics[width=4cm]{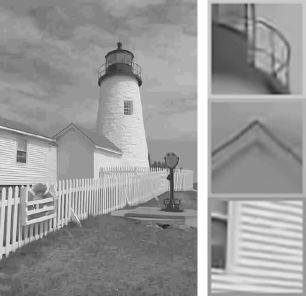}}\\
    \subfigure[ARCNN \cite{Dong2015Compression}]{
   \includegraphics[width=4cm]{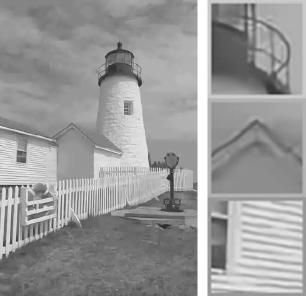}}
    \subfigure[TNRD \cite{chen2017trainable}]{
   \includegraphics[width=4cm]{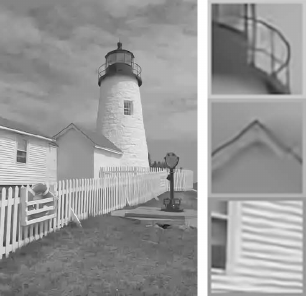}}
    \subfigure[DnCNN-3 \cite{Zhang2017Beyond}]{
    \includegraphics[width=4cm]{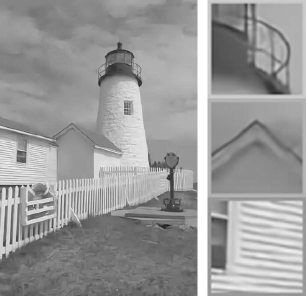}}
    \subfigure[Proposed DPW-SDNet]{
    \includegraphics[width=4cm]{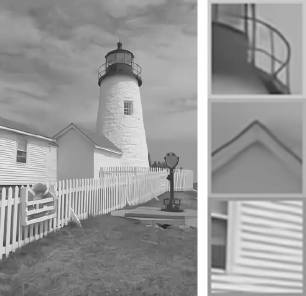}}
    \caption{Visual quality comparison of different soft decoding methods on \emph{Lighthouse3} in the case of QF = 10. (a) Original image (PSNR (dB), SSIM, PSNR-B (dB)); (b) JPEG (28.29, 0.7636, 25.98); (c) CONCOLOR \cite{Zhang2016CONCOLOR} (29.77, 0.7976, 29.36); (d) D2SD \cite{Liu2016Data} (29.77, 0.7977, 29.24); (e) ARCNN \cite{Dong2015Compression} (29.63,0.7973, 29.19); (f) TNRD \cite{chen2017trainable} (29.75, {\textcolor{blue}{0.8013}}, 29.27); (g) DnCNN-3 \cite{Zhang2017Beyond} ({\textcolor{blue}{29.81}}, 0.8007, {\textcolor{blue}{29.38}}); (h) Proposed DPW-SDNet ({\textcolor{red}{30.30}}, {\textcolor{red}{0.8104}}, {\textcolor{red}{29.76}}).}
    \label{fig.6}
\end{figure*}

\subsection{Soft Decoding Performance Evaluation}
The DPW-SDNet is compared with five state-of-the-art soft decoding algorithms for JPEG-compressed images, including two restoration-based approaches (i.e., CONCOLOR \cite{Zhang2016CONCOLOR} and D2SD \cite{Liu2016Data}) and three deep learning-based algorithms (i.e., ARCNN \cite{Dong2015Compression}, TNRD \cite{chen2017trainable}, and DnCNN-3~\cite{Zhang2017Beyond}). Referring to \cite{Zhang2017Beyond}, two benchmark imagesets Classic$5$ and LIVE$1$ are used as test datasets. For the color images in the LIVE$1$ dataset, only the luminance components are processed. The MATLAB JPEG encoder is used to generate JPEG-compressed images at different quality factors (QFs). We compare the performance of these algorithms in the cases of QF = $10$, $20$, $30$, and $40$. For the DPW-SDNet, a dedicated model is trained for each QF. For the five competitors, we use the original codes and models provided by the authors.

Table \ref{table.1} reports the objective assessment scores achieved by all tested algorithms, including the PSNR, SSIM \cite{wang2004image}, and PSNR-B \cite{yim2011quality} \footnote{\label{footnote2} For the TNRD \cite{chen2017trainable}, the results at QF = $40$ are not presented as the corresponding model is not available.}. Note that the PSNR-B is a specifically developed assessment metric for blocky and deblocked images. It can be observed from Table \ref{table.1} that the DPW-SDNet consistently outperforms the five competitors with considerable improvements. The only exception is the PSNR-B value on Classic$5$ in the case of QF = $40$, where the CONCOLOR \cite{Zhang2016CONCOLOR} is superior to the DPW-SDNet. Overall, the DnCNN-3 \cite{Zhang2017Beyond} and TNRD \cite{chen2017trainable} generate the second-best and the third-best results, respectively. The CONCOLOR~\cite{Zhang2016CONCOLOR}, D2SD \cite{Liu2016Data}, and  ARCNN \cite{Dong2015Compression} achieve comparable performance overall. On average, the proposed DPW-SDNet achieves about  ($0.30 \sim 0.34$) dB PSNR gains, ($0.0030 \sim 0.0098$) SSIM gains, and ($0.04 \sim 0.24$) dB PSNR-B gains over the second-best approach DnCNN-3~\cite{Zhang2017Beyond}. The gains over the two restoration-based soft decoding algorithms and ARCNN \cite{Dong2015Compression} are more significant. The improvements over state-of-the-art deblocking approaches demonstrate the effectiveness of the proposed DPW-SDNet.

One important aim of soft decoding algorithms is to recover images with high visual quality as JPEG-compressed images at high compression ratios usually suffer from severe artifacts. Therefore, some soft decoded images produced by different methods at QF = $10$ are shown in Fig.~\ref{fig.4}, Fig. \ref{fig.5}, and Fig. \ref{fig.6} in order to compare visual quality.  It can be observed that most of the compression artifacts in JPEG images  are removed by performing soft decoding on them. However, some soft decoded images are over-smoothed to some extent, or still suffer from visible artifacts. By contrast, the DPW-SDNet shows superiority in reducing artifacts and restoring details. The restored images using DPW-SDNet are more perceptually appealing, which can be seen from the highlighted regions. The results in this section verify that the DPW-SDNet not only achieves higher objective evaluation scores, but also produces better visual quality.
\begin{table*}[!tb]
\centering
\caption{Average PSNR (dB)/SSIM/PSNR-B (dB) scores of different variants of the DPW-SDNet on Classic$5$ and LIVE$1$. The best scores are highlighted in {\color{red}{red}}.}
\begin{tabular}{|p{1cm}<{\centering}|p{2.7cm}<{\centering}|p{2.75cm}<{\centering}|p{2.75cm}<{\centering}|p{2.75cm}<{\centering}|p{2.75cm}<{\centering}|}
\hline
\multicolumn{2}{|c|}{QF} & 10 & 20 & 30 & 40\\
\hline

\multirow{3}{*}{Classic5}
& P-SDNet  & 29.69/0.8116/29.33 & 31.89/0.8657/31.39 & 33.18/0.8899/32.49 & 34.04/0.9036/33.22\\
\cline{2-6}
& W-SDNet  & 29.70/0.8117/29.33 & 31.91/0.8660/31.37 & 33.18/0.8900/32.48 & 34.03/0.9036/33.21\\
\cline{2-6}
& DPW-SDNet  & {\textcolor{red}{29.74}}/{\textcolor{red}{0.8124}}/{\textcolor{red}{29.37}} & {\textcolor{red}{31.95}}/{\textcolor{red}{0.8663}}/{\textcolor{red}{31.42}} & {\textcolor{red}{33.22}}/{\textcolor{red}{0.8903}}/{\textcolor{red}{32.51}}& {\textcolor{red}{34.07}}/{\textcolor{red}{0.9039}}/{\textcolor{red}{33.24}}\\
\hline
\multicolumn{6}{|c|}{}\\
\hline
\multirow{3}{*}{LIVE1}
& P-SDNet  & 29.49/0.8203/29.10 & 31.86/0.8849/31.25 & 33.27/0.9126/32.49 & 34.26/0.9278/33.41\\
\cline{2-6}
& W-SDNet  & 29.51/0.8205/29.11 & 31.87/0.8850/31.25  & 33.28/0.9127/32.50 &  34.26/0.9279/33.42\\
\cline{2-6}
& DPW-SDNet  &  {\textcolor{red}{29.53}}/{\textcolor{red}{0.8210}}/{\textcolor{red}{29.13}}  & {\textcolor{red}{31.90}}/{\textcolor{red}{0.8854}}/{\textcolor{red}{31.27}}  & {\textcolor{red}{33.31}}/{\textcolor{red}{0.9130}}/{\textcolor{red}{32.52}} & {\textcolor{red}{34.30}}/{\textcolor{red}{0.9282}}/{\textcolor{red}{33.44}}\\
\hline

\end{tabular}
\label{table.2}
\end{table*}

\subsection{Discussion on Dual-Domain Soft Decoding}

In DPW-SDNet, two parallel branches are used to restore the compressed image in the pixel domain and wavelet domain, respectively. It is meaningful to study the ability of the two branches and discuss the effectiveness of the dual-domain combination. Table \ref{table.2} presents the objective assessment scores of the DPW-SDNet and its two variants, i.e., the P-SDNet and W-SDNet. Here the P-SDNet represents that only the pixel domain branch is used to restore the compressed image, while the W-SDNet represents that only the wavelet domain branch is used.

It can be observed from Table \ref{table.2} that both the P-SDNet and W-SDNet generate excellent restoration performance, which proves the ability of the presented network. Moreover, the gains of the DPW-SDNet over the P-SDNet and W-SDNet verify the effectiveness of the dual-domain soft decoding. Furthermore, it is believed that the fusion of the two branches could be more effective with a more complex combination method.

\begin{table*}[!tb]
\centering
\caption{Comparisons of PSNR (dB)/SSIM/PSNR-B (dB) scores of the DnCNN-3 \cite{Zhang2017Beyond}, DPW-SDNet, and B-DPW-SDNet on Classic$5$ and LIVE$1$. The best and the second-best scores are highlighted in {\color{red}{red}} and {\color{blue}{blue}}, respectively.}

\begin{tabular}{|p{1cm}<{\centering}|p{2.7cm}<{\centering}|p{2.75cm}<{\centering}|p{2.75cm}<{\centering}|p{2.75cm}<{\centering}|p{2.75cm}<{\centering}|}
\hline
\multicolumn{2}{|c|}{QF} & 10 & 20 & 30 & 40\\
\hline
\multirow{3}{*}{Classic5}
& DnCNN-3 \cite{Zhang2017Beyond}  & 29.40/0.8026/29.13 & 31.63/0.8610/31.19 & 32.90/0.8860/32.36 & 33.77/0.9003/{\textcolor{blue}{33.20}} \\
\cline{2-6}
& DPW-SDNet  & {\textcolor{red}{29.74}}/{\textcolor{red}{0.8124}}/{\textcolor{red}{29.37}} & {\textcolor{red}{31.95}}/{\textcolor{red}{0.8663}}/{\textcolor{red}{31.42}} & {\textcolor{red}{33.22}}/{\textcolor{red}{0.8903}}/{\textcolor{red}{32.51}}& {\textcolor{red}{34.07}}/{\textcolor{red}{0.9039}}/{\textcolor{red}{33.24}}\\
\cline{2-6}
& B-DPW-SDNet   & {\textcolor{blue}{29.69}}/{\textcolor{blue}{0.8104}}/{\textcolor{blue}{29.34}} & {\textcolor{blue}{31.92}}/{\textcolor{blue}{0.8660}}/{\textcolor{blue}{31.39}} & {\textcolor{blue}{33.18}}/{\textcolor{blue}{0.8900}}/{\textcolor{blue}{32.44}}& {\textcolor{blue}{34.01}}/{\textcolor{blue}{0.9035}}/33.19 \\
\hline
\multicolumn{6}{|c|}{}\\
\hline
\multirow{3}{*}{LIVE1}
& DnCNN-3 \cite{Zhang2017Beyond}  & 29.19/0.8123/28.91 & 31.59/0.8802/31.08 & 32.99/0.9090/32.35 & 33.96/0.9247/33.29 \\
\cline{2-6}
& DPW-SDNet  &  {\textcolor{red}{29.53}}/{\textcolor{red}{0.8210}}/{\textcolor{red}{29.13}}  & {\textcolor{red}{31.90}}/{\textcolor{red}{0.8854}}/{\textcolor{red}{31.27}}  & {\textcolor{red}{33.31}}/{\textcolor{red}{0.9130}}/{\textcolor{red}{32.52}} & {\textcolor{red}{34.30}}/{\textcolor{red}{0.9282}}/{\textcolor{red}{33.44}}\\
\cline{2-6}
& B-DPW-SDNet & {\textcolor{blue}{29.48}}/{\textcolor{blue}{0.8193}}/{\textcolor{blue}{29.10}} & {\textcolor{blue}{31.87}}/{\textcolor{blue}{0.8849}}/{\textcolor{blue}{31.26}} & {\textcolor{blue}{33.27}}/{\textcolor{blue}{0.9127}}/{\textcolor{blue}{32.46}}& {\textcolor{blue}{34.24}}/{\textcolor{blue}{0.9278}}/{\textcolor{blue}{33.38}}  \\
\hline
\end{tabular}
\label{table.3}
\end{table*}

\subsection{Discussion on Blind Soft Decoding}

In above experiments, we use a dedicated model for each compression QF. To test the capacity of the DPW-SDNet further, we train a universal model for compressed images at different QFs. We refer to the universal model as the blind DPW-SDNet (B-DPW-SDNet), which is trained using the samples compressed at different QFs \footnote{\label{footnote3} Note that the same training dataset and the same number of training samples are used to train the universal model and the dedicated model.}. In Section~4.2, DPW-SDNet and DnCNN-3 \cite{Zhang2017Beyond} perform the best and the second-best on the whole, respectively. Therefore,  we compare the B-DPW-SDNet with them in Table \ref{table.3}.

As expected, the B-DPW-SDNet is slightly inferior to DPW-SDNet. However, in most cases, it still outperforms DnCNN-3 \cite{Zhang2017Beyond} with obvious gains. Compared with DPW-SDNet, B-DPW-SDNet is more flexible and practical. Given QF, DPW-SDNet can be used to obtain better restoration performance, while B-DPW-SDNet can produce competitive results when the QF is unknown. Hence, one can select a proper model according to the practical application.

\begin{figure}[!tb]
    \centering
    \includegraphics[width=7cm]{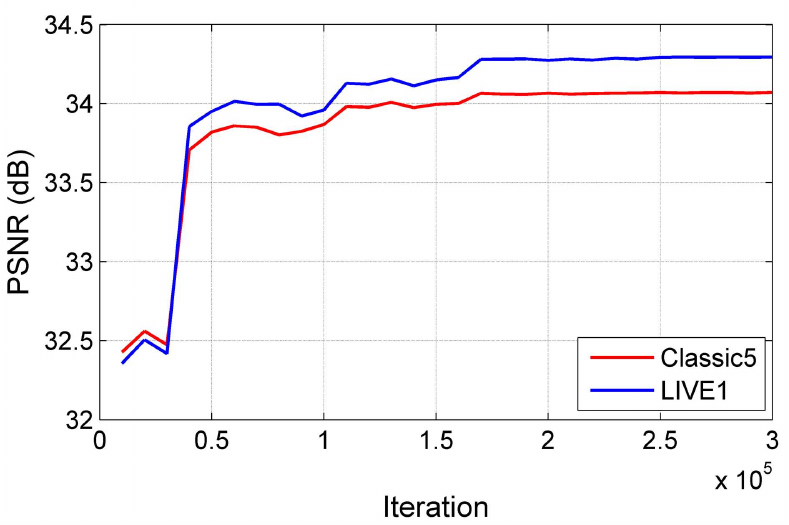}
    \caption{The PSNR (dB) values of DPW-SDNet on Classic$5$ and LIVE$1$ with different training iterations (QF = $40$). }
    \label{fig.7}
\end{figure}
\begin{figure}[!tb]
    \centering
    \includegraphics[width=7cm]{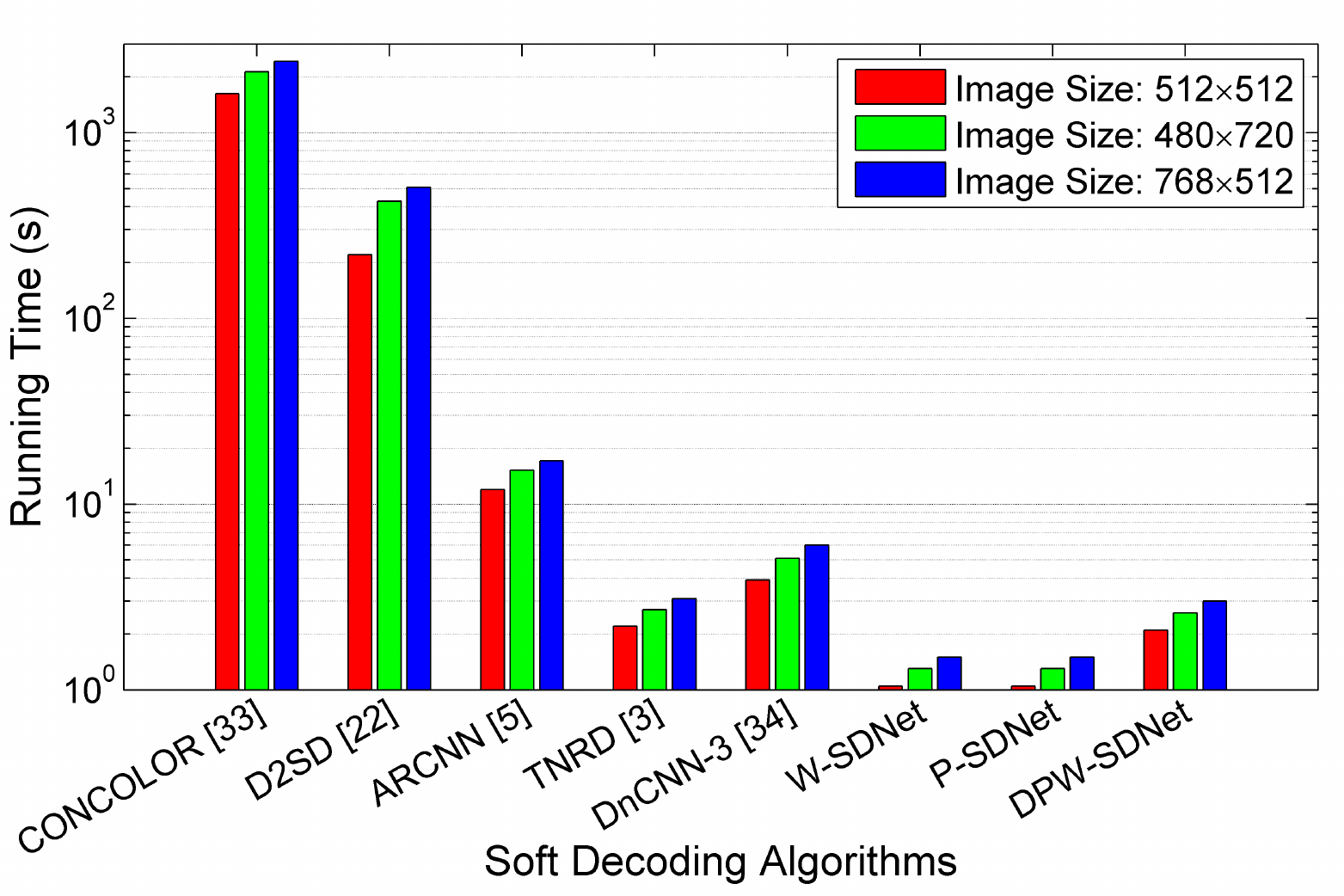}
    \caption{The running time (s) of different soft decoding algorithms on three representative image sizes in Classic$5$ and LIVE$1$. }
    \label{fig.8}
\end{figure}
\subsection{Empirical Study on Training Convergence and Running Time}

In Fig. \ref{fig.7}, we show the PSNR values of DPW-SDNet with different training iterations. The trends are similar for different QFs, so only the curves at QF = $40$ are presented. It can be seen that the training converges after about 200,000 iterations. In our experiments, the maximum number of iterations is set to 300,000. The training of a single model takes about $9$ hours on a GeForce GTX $1080$ Ti GPU.

Running time is an important factor for a soft decoding algorithm. We run different deblocking methods on the same desktop computer with an Inter Core i$7$ CPU $4.2$ GHz, 32GB RAM, and Matlab environment. Fig. \ref{fig.8} presents the execution time of different approaches on three representative image sizes in Classic$5$ and LIVE$1$ \footnote{\label{footnote4} In this experiment, the running time of the TNRD \cite{chen2017trainable} is evaluated with the multi-threaded computation implementation.}. It can be seen that the proposed P-SDNet and W-SDNet are the most efficient approaches. The DPW-SDNet costs about $2\times$ time compared with the P-SDNet and W-SDNet, but it is still less time-consuming than other compared algorithms. Moreover, the execution speed of the DPW-SDNet can be greatly accelerated with a GPU.

\section{Conclusion and Future Work}
A dual pixel-wavelet domain deep network-based soft decoding framework is developed for JPEG-compressed images, namely DPW-SDNet. In DPW-SDNet, the compressed image is restored in both pixel and wavelet spaces using deep CNNs. In addition, we use 4-channel tensors as the inputs of our networks rather than the 2-dimensional images, which makes the DPW-SDNet efficient and effective. Experimental results on benchmark datasets demonstrate the effectiveness and efficiency of our soft decoding algorithm. Future work includes the extensions of the proposed DPW-SDNet to other image compression standards as well as other image restoration problems.

\section{Acknowledgment}
This work was supported in part by the National Natural Science Foundation of China under Grant 61471248, in part by the Fundamental Research Funds for the Central Universities under Grant 2012017yjsy159, and in part by the China Scholarship Council under Grant 201706240037. The authors thank Cheolhong An and Wenshu Zhan for helpful discussions.

\newpage

{\small
\bibliographystyle{ieee}
\bibliography{Paper_Deblocking}
}

\end{document}